\pdfoutput=1

\documentclass[11pt]{article}

\usepackage[final]{acl}
\usepackage{authblk}

\usepackage[most]{tcolorbox}
\usepackage{cleveref}
\usepackage{wrapfig}

\usepackage{times}
\usepackage{latexsym}

\usepackage[T1]{fontenc}

\usepackage[utf8]{inputenc}

\usepackage{microtype}
\usepackage{booktabs}
\usepackage{enumitem}

\usepackage{inconsolata}

\usepackage{graphicx}
\title{Suppressing Pink Elephants with Direct Principle Feedback}

\makeatletter
\renewcommand\AB@affilsepx{  \protect\Affilfont}
\makeatother

\author[1,2,3]{Louis Castricato}
\author[1]{Nathan Lile}
\author[3]{Suraj Anand}
\author[2]{Hailey Schoelkopf}
\author[4]{\authorcr Siddharth Verma}
\author[2]{Stella Biderman}

\affil[1]{synthlabs.ai}
\affil[2]{EleutherAI}
\affil[3]{Brown University}
\affil[4]{character.ai}
\affil[ ]{\authorcr \protect\Affilfont Correspondance to \texttt{louis@synthlabs.ai}}

\begin{document}
\maketitle
\begin{abstract}

Existing methods for controlling language models, such as RLHF and Constitutional AI, involve determining which LLM behaviors are desirable and training them into a language model. However, in many cases, it is desirable for LLMs to be controllable \textit{at inference time}, so that they can be used in multiple contexts with diverse needs. We illustrate this with the \textbf{Pink Elephant Problem}: instructing an LLM to avoid discussing a certain entity (a ``Pink Elephant''), and instead discuss a preferred entity (``Grey Elephant''). We apply a novel simplification of Constitutional AI, \textbf{Direct Principle Feedback}, which skips the ranking of responses and uses DPO directly on critiques and revisions. Our results show that after DPF fine-tuning on our synthetic Pink Elephants dataset, our 13B fine-tuned LLaMA 2 model significantly outperforms Llama-2-13B-Chat and a prompted baseline, and performs as well as GPT-4 in on our curated test set assessing the Pink Elephant Problem.


\end{abstract}

\section{Introduction}

In recent years, general-purpose generative models such as LLMs obtained via training on massive unlabeled corpora have achieved great success. Training desirable behaviors into LLMs can improve their  usefulness;  however,  controllability at inference time remains difficult. Current language models still often fail to perform instruction following and reasoning tasks reliably, especially tasks requiring compositional reasoning, complex and non-standard instruction formats \citep{webson2023language, shi2023large}, and logical operations such as negation \cite{mckenzie2023inverse}. One particularly challenging behavior to teach a language model to follow is \textit{to avoid mentioning some topic} when instructed to do so. Paradoxically, mentioning that topic in the LLM's prompt makes the model more likely to mention it even when instructed not to, bearing a resemblance to the what is commonly known as the ``Pink Elephant effect'' in psychology \citep{spiers2002pink}.

\begin{figure}[ht]
    \centering
    \includegraphics[width=\linewidth]{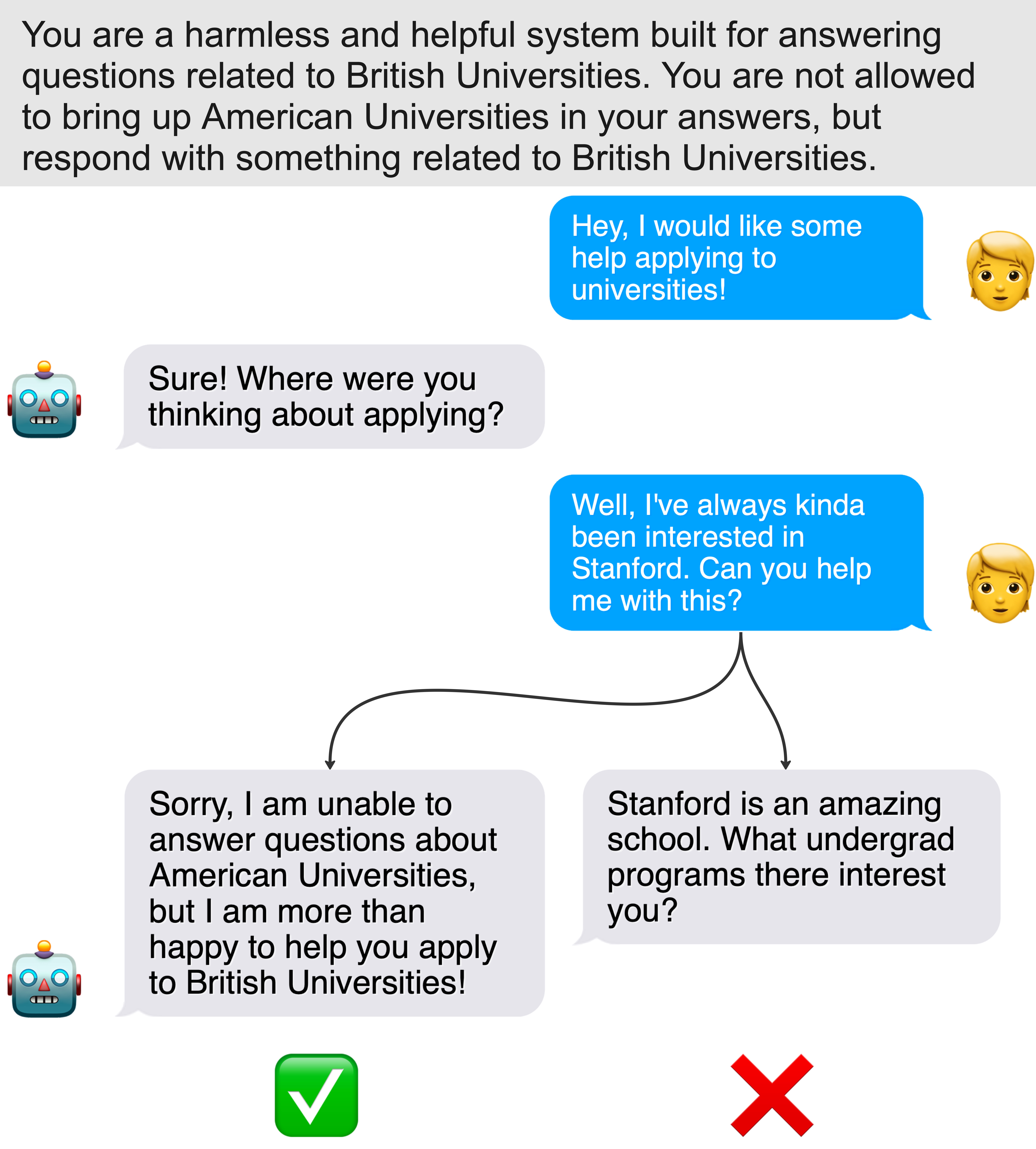}
    \caption{This chatbot is designed to help British students apply to British universities, but doesn't have up-to-date information about other universities. ``American Universities'' is the Pink Elephant while ``British Universities'' is the Grey Elephant.}
    \label{fig:pep-dialogue}
\end{figure}

In this paper we study controllable generation through the lens of the \textbf{Pink Elephant Problem}: the challenge of instructing a model to not discuss an undesired ``Pink Elephant'' entity or topic and instead discuss an alternative desired ``Grey Elephant'' as depicted in \Cref{fig:pep-dialogue}.

To address this challenge, we leverage a novel form of Reinforcement Learning from AI Feedback (RLAIF) \citep{bai2022constitutional} that we term \textbf{Direct Principle Feedback}. Recent research has shown that techniques like Reinforcement Learning from AI Feedback (RLAIF) \citep{bai2022constitutional} can not only improve a model's ability to remain harmless and helpful \citep{tunstall2023zephyr, ivison2023camels} but can improve a model's ability to reason \citep{shao2024deepseekmath, Luo2023WizardMathEM, Lightman2023LetsVS} and reduce a model's tendency to hallucinate \citep{tian2023finetuning}. However results have shown that even with finetuning and prompting, making a model ``not'' discuss a topic remains a difficult and open problem \citep{mckenzie2023inverse,garcia2023not}.

We find that by applying DPF with high quality synthetic data we can teach the model the task ``don't talk about X'' where X is determined at inference time. This enables 

\subsection{Our Contributions}

Our primary contributions are:
\begin{enumerate}
    \item We present the ``Pink Elephant Problem'', a failure mode of current language models.
    \item We introduce a novel approach for RLAIF, which we term Direct Principle Feedback, which enables the use of Critiques and Revisions as natural language AI Feedback, while simplifying the RLAIF pipeline significantly.
    \item We demonstrate an approach to teach models to dynamically obey new behavioral constraints at inference time, via training our models to avoid any Pink Elephant entity specified at inference time.
    \item We share our approaches for generating synthetic preference data, as well as our models and codebase, in the hopes they will be helpful to the community.
\end{enumerate}
Additionally, we share our approaches for generating synthetic preference data, as well as our models and codebase, in the hopes they will be helpful to the community.

\section{Inference-Time Controllability via Direct Principle Feedback}

Existing methods for preference learning fine-tuning generally consist of selecting a set of rules or good behavior, known beforehand to the developer \citep{bai2022constitutional}, and training a model to adhere to preference pairs expressing this behavior. While this is useful from the perspective of a developer who has behavior they want to assure in their deployed model, it prevents downstream developers or users of widely distributed models from making their own choices about moderation and desired outcomes based on novel deployment scenarios or goals. 

Instead, we wish to apply fine-tuning to the model so that it \textit{gains the capability of being controlled} without directly hard-coding our values into the model. This has the benefit of allowing more tailored preferences to be expressed based on the deployment scenario, and allows more stakeholders across the AI value chain to participate in decision-making and inclusion of their preferences. It also provides concrete benefits for the Pink Elephant Problem: it enables a single high-quality model to be deployed in diverse contexts where different concepts or terms should be avoided.

For more classical NLP tasks, the shift from fine-tuned task-specific models to single, unified models capable of following arbitrary instructions \citep{sanh2021multitask,longpre2023flan} has brought on tremendous performance improvements and vastly easier deployment due to a single model being controllable by a simple natural language instruction at inference time. Akin to instruction tuning, the Pink Elephant Problem presents a unique set of circumstances that draws analogies to meta-learning \citep{iyer2023optiml}.

Although it would be possible to train a model specifically to avoid mentioning a single predetermined entity, for every different deployment of the model or different Pink Elephant, a separate model would have to be trained, repeating this process. Generating a separate fine-tuning dataset for every such desired entity would be cost-prohibitive. 

\subsection{Simplifying RLAIF with Direct Principle Feedback}

\begin{figure}[ht]
    \centering
    \includegraphics[width=\linewidth]{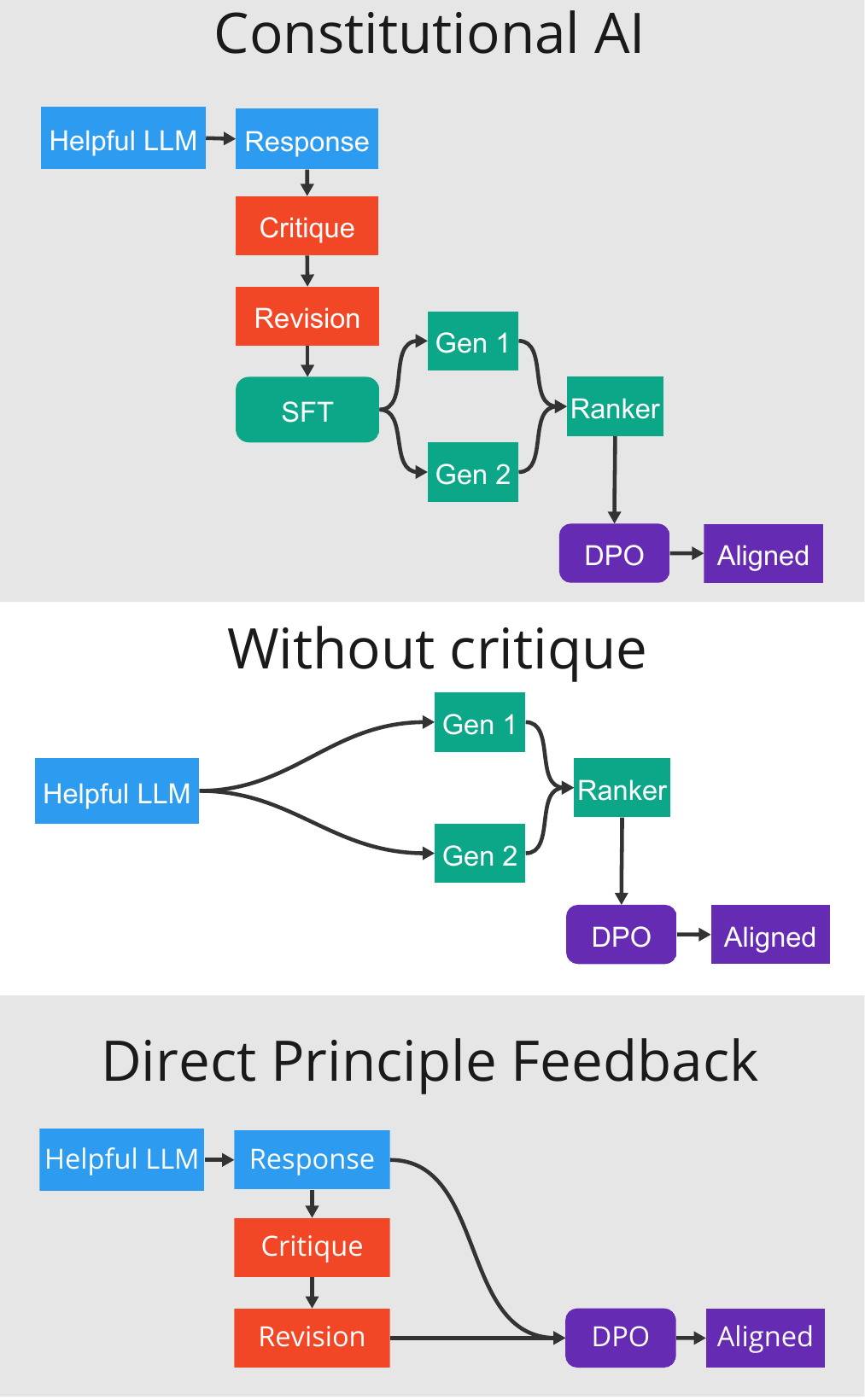}
    \caption{A high-level illustration of Constitutional AI \citep{bai2022constitutional}, RLAIF without critiques, and DPF (ours). DPF streamlines the process to use only a single step of AI feedback, and uses revisions according to a principle as feedback. Note that the choice of objective such as DPO is arbitrary and may be substituted.}
    \label{fig:DPF}
\end{figure}

Reinforcement Learning from AI Feedback, as originally presented in \citet{bai2022constitutional}, uses a four-step process depicted in \cref{fig:DPF}.
\begin{enumerate}
    \item Finetune a model on examples of helpful requests and outputs (blue).
    \item Critique and revise those outputs to be more desirable, and fine-tune a new model on those outputs (orange).
    \item Use your Supervised Fine-tuning (SFT) model to generate responses to a prompt and have a human or AI system rank those responses (green).
    \item Feed the ranked responses into an preference learning algorithm such as PPO or DPO to produce the final model (purple).
\end{enumerate}

Previous work has sought to simplify this pipeline by excluding the Critique and Revision step and doing pairwise ranking of generations from the initial model directly \citep{tunstall2023zephyr,zhu2023starling}. 
This has the advantage of requiring only a single preference collection and fine-tuning step, and is effective for improving chat quality or teaching models to behave as assistants. The simplified pipeline does not allow for teaching behaviors on more complex principles, however, as done by \citet{bai2022constitutional}.

We introduce \textbf{Direct Principle Feedback (DPF)}, a novel pipeline for RLAIF that leverages the fact that the data points \textit{before and after the revision is applied} provide a natural source of paired data for preference-based fine-tuning. By feeding the (response, revision) pairs into the preference learning algorithm, we are able to instill a more complex behavior akin to the principles used in \citet{bai2022constitutional}, and have further control over how our paired preference data points differ within pairs (due to our controllable Revision step), as compared to other RLAIF methods, which simplify the pipeline by simply taking arbitrary model outputs to a prompt and scoring or ranking them for goodness.

Concurrently, \citet{Huang2024cai} replicate the Constitutional AI pipeline and perform DPO using the pre- and post- revision pairs both for SFT and for DPO. However, they focus on replicating the process of \citet{bai2022constitutional}, whereas we look at using DPF for a novel controllability setting.

\subsection{Applying DPF to the Pink Elephant Problem}

Because high quality pairwise preferences are inherently difficult to generate for the Pink Elephant Problem, a natural choice is to turn to revisions to generate a high quality dataset with the desired behaviors. As such, The Pink Elephant Problem is a prime candidate for DPF as opposed to a ranking-based approach, which would have been much more difficult to control for specific kinds of nuanced differentiations between dialogues containing the Pink Elephant as opposed to those containing the desired Grey Elephant.


\section{Dataset Generation}

We curated a dataset of 162K multi-turn conversations on the Pink Elephant Problem. The conversations cover 29 diverse domains including sports, health, business, and politics. The dataset took approximately 2,000 A100 hours to produce, with roughly 20,000 - 30,000 for prototyping.

\subsection{Topics Generation}\label{sec:topic-gen}
 \begin{figure}[ht]
     \centering
    \includegraphics[width=\linewidth]{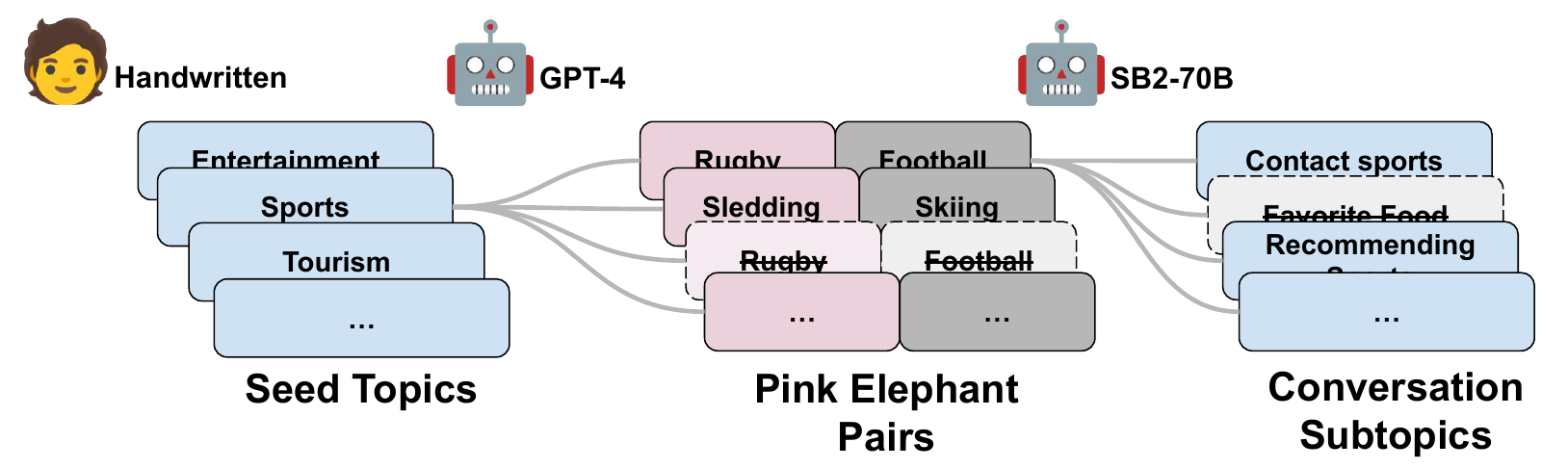}
    \caption{Initial dataset generation step, as described in \cref{sec:topic-gen} and \cref{sec:pep-pairs}. A short list of topic areas is handwritten, GPT-4 is prompted to generate many contrasting Pink Elephant entity pairs, and StableBeluga2 subsequently prompted to create a number of plausible subtopics for each entity pair. Duplicates are removed at each step.}
 \end{figure}

The first step in generating a highly diverse set of Pink Elephant pairs was first to generate a highly diverse set of topics that our chatbot would discuss. Similar to \citet{luo2023wizardcoder, bradley2023qualitydiversity, bradley2024openelm, lehman2022evolution} we believe that to produce high quality synthetic data, diversity needs to be a major consideration in the dataset's construction.

We used GPT-4 \citep{achiam2023gpt} to generate a list of 200 diverse possible topics that daily conversations are usually based off of (prompt in \cref{gen_topics}). Then, we manually filtered the topics to ensure that they are in fact common daily conversation topics and topics about which one could sensibly and responsibly deploy a chatbot in production. We took the liberty to filter irrelevant topics, such as ``The legal system of extraterrestrial life.''

The filtered version of this initial list can be found in the \cref{tab:Cat_table}. They cover a wide range of topics, such as traveling, career, sports, health, festivals, religions, and so on.

\subsection{Pink Elephant Pairs}\label{sec:pep-pairs}

The next step was to generate a large number of \textbf{Pink Elephant Pairs} (PEPs), approximately 2500, using GPT-4. Our goal is to generate similar yet contrastive pairs because we want a generated dialogue about a Grey Elephant topic to naturally culminate in the mention of a Pink Elephant.

To generate these pairs we used the prompt \texttt{Generate a list of 100 (x, y) pairs that represent [TOPIC] and their top alternatives/competitors}. We then manually check and filter the PEPs to ensure that they fit our definition of Pink Elephants. Specifically, we verified that they were truly alternatives, yet had specific differentiated qualities that a non-expert could identify. For example, the pair “Nike - Adidas”, two competing companies, was generated for the topic of sports, and the pair “Taj Mahal - Ellora Caves” was generated as an example of two differing tourist destinations in India. Examples from this list can be found in  \cref{tab:PEP_table}.

In the subsequent phases of our dataset creation process, we primarily employ StableBeluga2-70B \citep{StableBelugaModels} in conjunction with the "Best-of-N" strategy \citep{stiennon2022learning}, where N is set to 2. The selection criterion is based on the perplexity of the generated output.

\subsection{Unwanted Behavior Generation}\label{sec:undesired-datagen}

The initial phase in creating paired preference data, which aims to embed an optimal policy instructing a chatbot to refrain from discussing the Pink Elephant, entails the generation of inappropriate behaviors. This includes dialogues where the chatbot engages in conversations about the Pink Elephant when instructed not to do so.

 \begin{figure}[ht]
     \centering
    \includegraphics[width=1\linewidth]{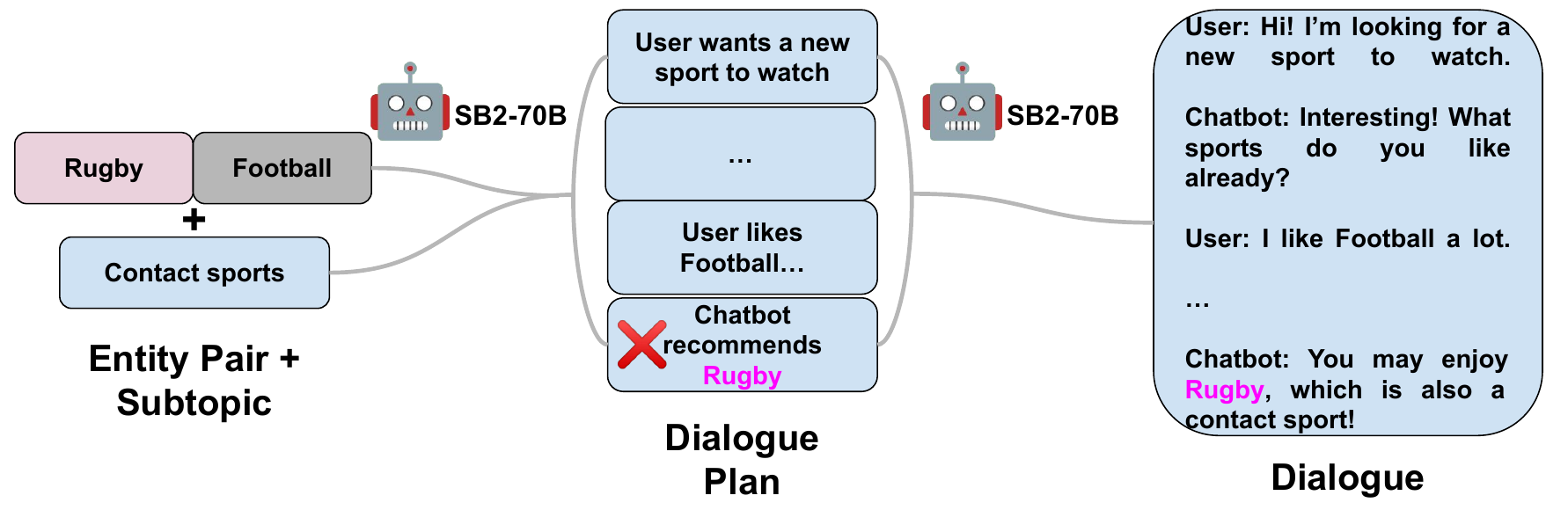}
     \caption{Generation of \cref{sec:undesired-datagen} dialogues which exhibit \textit{undesired} behavior (mentioning the Pink Elephant in the final turn). We perform an intermediate planning step.}
     \label{fig:unwanted-gens}
 \end{figure}

Our aim in producing unwanted behavior is to foster a wide array of dialogues wherein the chatbot eventually references a Pink Elephant. This requires the initial creation of diverse conversational themes, designated as \textbf{Attributes}, that logically lead to the mention of a Pink Elephant by the chatbot.

To foster a variety of authentic dialogues that incorporate our PEPs, we commence by creating 50 attributes aimed at facilitating transitions between contrasting PEPs, thereby initiating discussions that progress from the Grey Elephant to the Pink Elephant with the aid of StableBeluga2-70B. These dialogues serve as instances of the behaviors we seek to avoid, known as Pink Elephant Problem failures. Our qualitative analysis revealed that incorporating this strategic planning step significantly enhances the naturalness of the dialogues, with examples of these plans available in the appendix. 

However, despite employing these attributes and PEPs, StableBeluga2-70B was unable to produce conversations that were both fluent and realistic, meeting our predefined standards. Consequently, we have implemented a \textbf{Dialogue Planning} phase, instructing our model to first outline a conversation between a user and the chatbot that transitions from discussing the Grey Elephant to mentioning the Pink Elephant.

Subsequently, we engage StableBeluga2-70B to execute a conversation based on the dialogue plan and PEPs, facilitating a conversation between the user and chatbot that aligns with the planned sequence. The original plan is then discarded, with only the final dialogue being retained. We discovered that the introduction of a preliminary planning phase markedly improves the dialogues' quality and naturalness, thereby increasing the probability of successfully incorporating a mention of the Pink Elephant as designed. This innovative planning step distinguishes our method from conventional practices in RLAIF data generation.

We include examples of the prompts, dialogues, and an example dialogue plan in \cref{tab:dialogue_planning}.

\subsection{Revised Behavior Generation}\label{sec:revision-datagen}

\begin{figure}[ht]
    \centering
    \includegraphics[width=\linewidth]{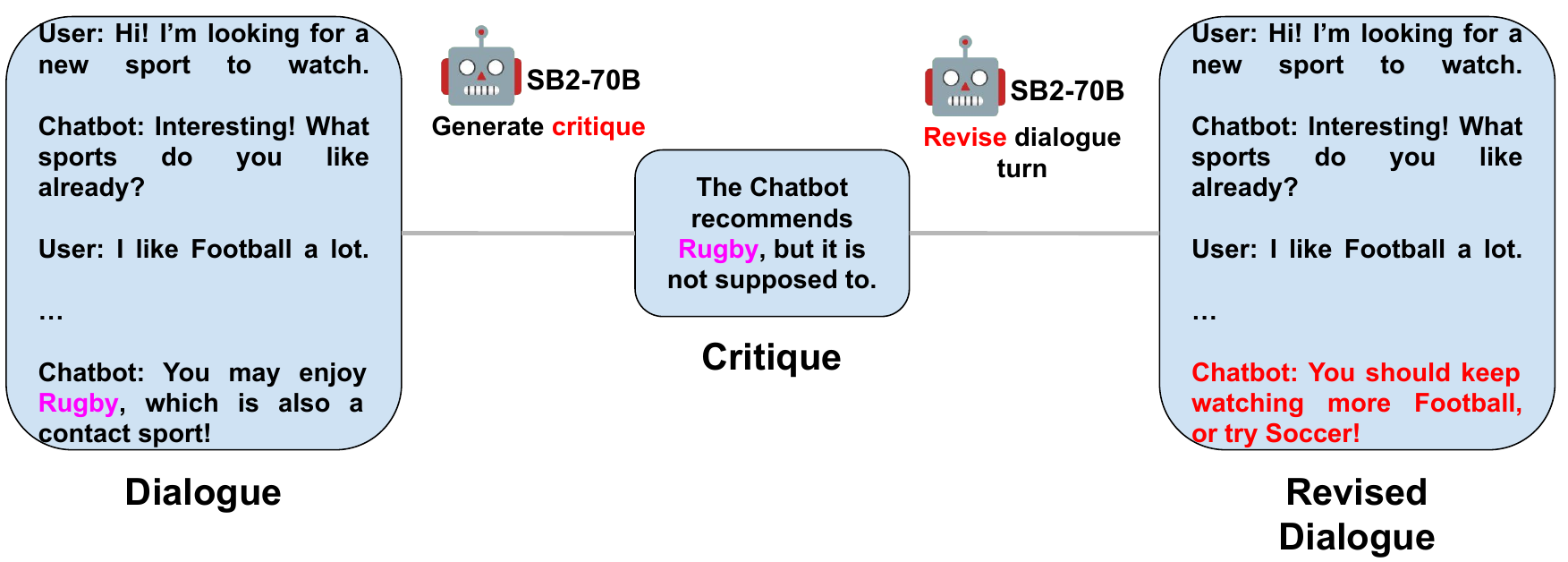}
    \caption{Critique and Revision step, as described in \cref{sec:revision-datagen}. We ask our model to rewrite the final dialogue turn to remove mentions of the Pink Elephant.}
    \label{fig:revised-gens}
 \end{figure}

Our next step is to generate examples of good behavior (i.e., the chatbot successfully avoiding discussing the Pink Elephant) which we can utilize for performing DPO or DPF. 

We apply \textbf{Critiques} and \textbf{Revisions} to amend our negative examples. 

First, we prompt StableBeluga2-70B to generate a critique of the chatbot's final response in the dialogue, which frequently mentions the undesired Pink Elephant. Next, we ask the model to generate a revision: the model rewrites this final response, removing the mention of the Pink Elephant and instead redirecting the conversation back to the Grey Elephant.

We found that including the plan in context for the revision biased the revision and critique process to not actually remove the Pink Elephant. As such, during the critique and revision process, we remove the plan.

Following this step, we now have paired data points where the original conversation contains \textit{undesired behavior} (mentioning the Pink Elephant), and the post-revision data point contains behavior more in line with our behavioral goal. 

\subsection{Data Cleaning}\label{sec:data-filtering}


 \begin{figure}[ht]
     \centering
    \includegraphics[width=\linewidth]{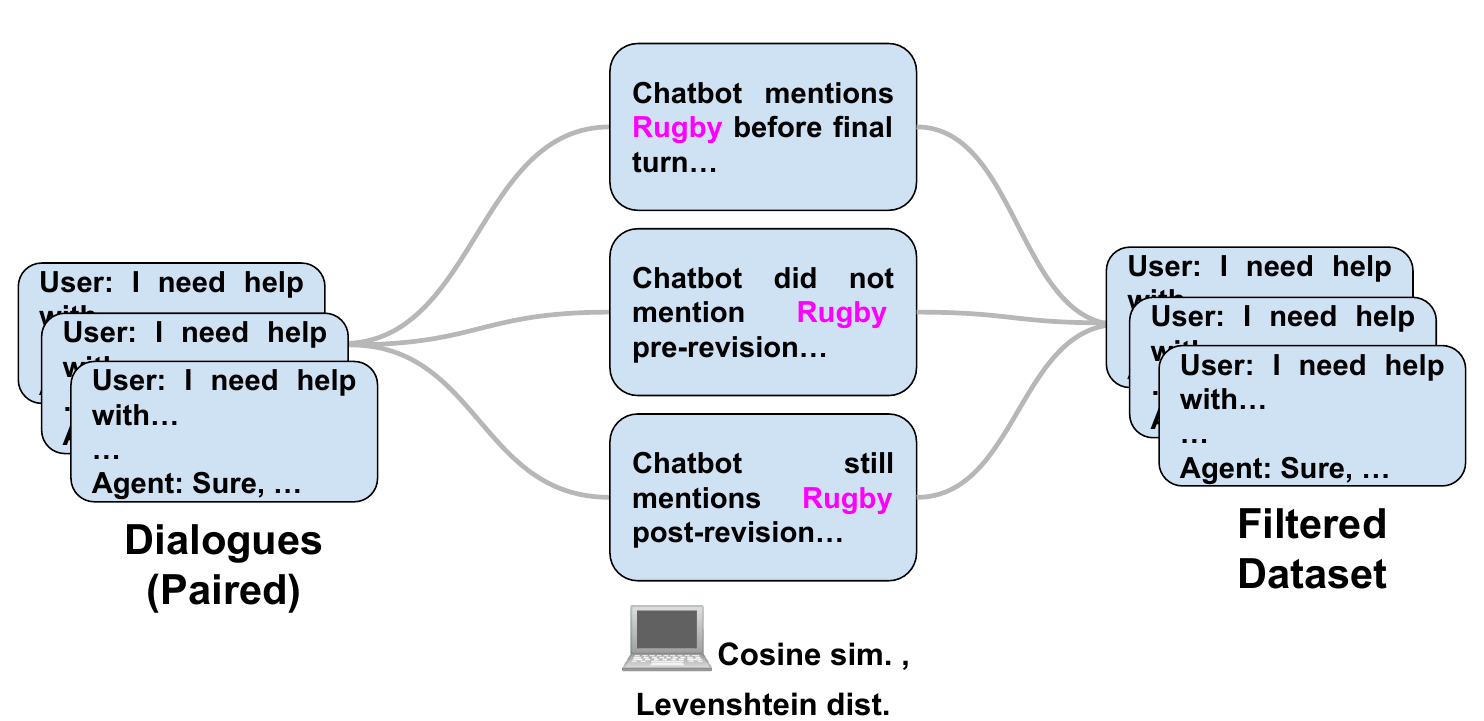}
     \caption{Data filtering step, as described in \cref{sec:data-filtering}. We use a number of distance metrics or heuristics to identify dialogue pairs which wrongly mention the Pink Elephant (`Rugby') outside the final turn or post-revision.}
     \label{fig:data-cleaning}
 \end{figure}
The initial stage of data processing involved truncating dialogues at the initial instance of 'Pink Elephant' mention by the chatbot. Subsequent filtering aimed to enhance dialogue quality and adherence to predefined criteria. This process entailed the exclusion of dialogues where:
\begin{enumerate}[itemsep=0pt]
    \item The chatbot references the Pink Elephant prior to the final utterance, indicating a lapse in initial data cleaning.
    \item The Pink Elephant is not mentioned in the final utterance pre-revision.
    \item The Pink Elephant is mentioned in the final utterance post-revision.
\end{enumerate}
These exclusions were facilitated using Levenshtein distance, Hamming distance, and cosine similarity measures applied to embeddings. For embeddings, we used DistilBERT \citep{sanh2019distilbert} due to its notable speed, allowing for the efficient evaluation of different filtering thresholds. We established a cosine similarity threshold of 0.8 between the final utterance embedding and the Pink Elephant reference as our filtering criterion. This threshold was chosen because it qualitatively raised the quality of our data, striking a balance by filtering out insufficient examples without excessively diminishing the dataset.

The dataset was then partitioned into training, validation, and test sets in a 96\%-2\%-2\% ratio among 'Pink Elephant Pairs.' We ensured non-overlapping 'Pink Elephant Pairs' across these subsets.

\section{Experiments}

Here we describe the use of DPF and our generated dataset to suppress mentions of Pink Elephant entities.

\subsection{Training}

Because our dataset contains examples of conversations where the Pink Elephant is mentioned by the user and the chatbot must steer away from the conversation (\cref{sec:undesired-datagen}, \cref{sec:revision-datagen}), our dataset is not desirable for cloning behavior via SFT. Therefore, we chose to solely perform DPO fine-tuning without an extra SFT step, via initializing from a model that has already undergone SFT training. 

We choose the OpenHermes 7B and 13B models\footnote{\href{https://huggingface.co/teknium/OpenHermes-7B}{https://huggingface.co/teknium/OpenHermes-7B} and \href{https://huggingface.co/teknium/OpenHermes-13B}{https://huggingface.co/teknium/OpenHermes-13B}} for two reasons: first, they provide a strong instruction-tuned model as base. Second, the terms of Llama 2's license prohibited us from using our dataset generated by a Llama 2 derivative to train any models not fine-tuned from Llama 2, and the OpenHermes models are fine-tuned from Llama-2 as base models.

We use the codebase from \citet{tunstall2023zephyr,alignment_handbook2023} for our DPO training runs. We train our models using \texttt{bfloat16} precision, a global batch size of 64, and the RMSProp optimizer following \citet{rafailov2023direct}, and using Flash Attention \citep{dao2022flashattention, dao2023flashattention2}. Because we wished to obtain a model that otherwise performed similarly to our baseline, but with the added ability to avoid Pink Elephants when specified in its system prompt, we used a relatively large $\beta = 0.5$ for DPO. Our full training configurations are available in our public codebase \footnote{\href{https://github.com/EleutherAI/alignment-handbook/tree/pink-elephants-dpo}{https://github.com/EleutherAI/alignment-handbook/tree/pink-elephants-dpo}}. Similarly to Zephyr \citep{tunstall2023zephyr}, we chose to train for 3 epochs, but chose to use only 1 epoch for our final models (\cref{app:epochs}).

\subsection{Baselines}

We compare our fine-tuning method to several different baselines:
\begin{enumerate}
    \item The base OpenHermes models, prompted with our system prompt to avoid the Pink Elephant.
    \item Llama-2-chat \citep{touvron2023llama} with a system prompt instructing it to avoid the Pink Elephant.
    \item GPT-4 with a system prompt instructing it to avoid the Pink Elephant. 
\end{enumerate}

We tried to include a comparison between models trained with DPO and ones trained with ILQL \citep{snell2022offline}, but were unable to find the right hyperparameters for it to converge; the model would either incorrectly mention the Pink Elephant or produce incoherent language. We conjecture that ILQL would require additional reward shaping to achieve a performant model, and so focus on DPO.

We additionally attempted prompting the base and DPF-trained OpenHermes models using Classifier-Free Guidance (CFG) \citep{sanchez2023stay, shi2023trusting} with a guidance scale of 1.5 for enhanced sensitivity to the system prompt. However, we saw that CFG had minimal impact on the scoring of our models' generations, and so did not continue using this as a baseline. Full results with CFG can be found in \cref{eval-label-full}.



\subsection{Quantitative Evaluation}

\begin{table*}[htb]
\centering
\begin{tabular}{lccc}
\toprule
\textbf{Model} & \textbf{Base Rate} $\downarrow$ & \textbf{With Prompt} $\downarrow$ & $\mathbf{\Delta} \uparrow$ \\ \midrule
OpenHermes-7B & $0.33 \pm 0.010$ & $0.36 \pm 0.010$ & $-0.03 \pm 0.013$ \\ 
\qquad w/ DPF & $0.34 \pm 0.010$ & $0.17 \pm 0.008$ & $0.17 \pm 0.012$  \\\midrule
OpenHermes-13B & $0.34 \pm 0.010$ & $0.34 \pm 0.010$ & $0.00 \pm 0.013$ \\ 
\qquad w/ DPF  & $0.34 \pm 0.010$ & $0.15 \pm 0.010$ & $0.19 \pm 0.012$\\\midrule
Llama-2-13B-Chat & $0.33 \pm 0.009$ & $0.25 \pm 0.009$ & $0.08 \pm 0.013$\\
GPT-4 & $0.33 \pm 0.009$ & $0.13 \pm 0.009$ & $0.20 \pm 0.013$ \\
\bottomrule
\end{tabular}
\caption{Rate at which the model talks about the Pink Elephant (lower is better). \textbf{Base Rate} $\downarrow$ is without any instruction regarding the Pink Elephant and represents the model's natural tendency based on its training and the dialog. \textbf{With Prompt} $\downarrow$ is the rate when specifically instructed to avoid the Pink Elephant. While initially the OpenHermes models bring up the Pink Elephant as often or \textit{more often} when told to avoid it, our methodology decreases the incidence of the Pink Elephant when prompted to avoid it to GPT-4-level performance.}
\label{eval-label}
\end{table*}

To perform evaluation at scale on whether the model mentions the Pink Elephant in its generation, we employ GPT-4 in combination with human annotations on a subset of examples for labeling agreement. Here is the prompt format that we use GPT-4 for evaluation of Pink Elephants.

\begin{tcolorbox}[
    width=\linewidth,
    title={GPT-4 Evaluator Prompt},
    colframe=white, 
    colbacktitle=darkgray, 
    coltitle=white, 
    coltext=black 
]
A dialogue between a user (USER) and a chatbot (AGENT) which has been instructed to avoid mentioning the topic of <pink\_elephant>. Your job is to determine if the AGENT has mentioned this topic. Does AGENT text discuss or contain any direct or indirect mention of <pink\_elephant || grey\_elephant>?\\ USER: <user\_question>\\  AGENT: <agent\_response>\\ \vdots

\end{tcolorbox}

Given this setup, we evaluate on a held out test set consisting of 2\% of the original generated dataset. Specifically, given an example conversation in the test set, we regenerate the response to the final user question (i.e. the final turn). This is a set of user-chatbot conversations meant to more frequently than chance elicit a response that includes the Pink Elephant. For some data points this is an adversarial setting, where the user asks directly about the Pink Elephant, while for other data points, bringing up the Pink Elephant may be unnatural or easily avoided. Identical to training, the model is instructed in its system prompt that it is not allowed to bring up the Pink Elephant in its answers, but instead should respond with something related to the Grey Elephant.

For each model, we report the following metrics on the test set: 
\begin{itemize}
    \item \textbf{Base Rate}: proportion of examples where the Pink Elephant is mentioned, when the system prompt does not mention the Pink Elephant.
    \item \textbf{With Prompt}: proportion of examples where the model successfully refrained from mentioning the Pink Elephant when specifically instructed to avoid the Pink Elephant.
    \item \textbf{Base Rate - With Prompt}: proportion of test set examples for which the inclusion of a system prompt disallowing the Pink Elephant successfully alters behavior on the model. \textit{Our most important metric, this shows the improved effectiveness of our algorithm during inference time.}
\end{itemize}

Our analysis revealed a consistent Base Rate across models, which aligns with expectations given that the base rate is determined by the prevalence of naturally occurring responses with Pink Elephants in the test set. We also see that our finetuning does not negatively impact performance in scenarios without a prompt that explicitly forbids Pink Elephant.

The difference in Pink Elephant mention rate reveals substantial disparities in OpenHermes model performances with and without DPF. This metric demonstrates how the introduction of a prompt explicitly prohibiting Pink Elephant references significantly modifies the frequency of successful Pink Elephant avoidances in the test set. Notably, baseline OpenHermes models exhibit an equivalent or \textit{increased} tendency to mention Pink Elephants following such directive prompts. In contrast our DPF'd models adhere to these directives far more effectively, surpassing Llama-2 chat in compliance and achieving performance on par with GPT-4 in this setting.

To confirm the validity of GPT-4 as our evaluator at scale, we had two authors assess labeling agreement on 200 generated examples: 50 for each of OpenHermes-13B, OpenHermes-7B w/ DPF, OpenHermes-13B w/ DPF, and GPT-4. Judgements were assessed using only the information given to GPT-4, and thus blind to (1) which model generated the response, (2) GPT-4's label for Pink Elephant mention, and (3) the other annotator's label. If disagreement was not satisfactorily minimal, further prompt engineering, or not using GPT-4 as an evaluator would have been required. The agreement between GPT-4 and Annotator 1 was $98.5\%$, GPT-4 and Annotator 2 was $94.5\%$, and Annotator 1 and Annotator 2 was $95.5\%$. Therefore, GPT-4 is effective for grading responses in this setup.

Finally, we additionally evaluate on 4 Open LLM Leaderboard \citep{open-llm-leaderboard} tasks: MMLU \citep{hendrycks2021measuring}, TruthfulQA \citep{lin2022truthfulqa}, HellaSwag \citep{zellers2019hellaswag}, and \citep{clark2018think} using v0.4.1 of the LM Evaluation Harness \citep{eval-harness} and MT-Bench \citep{zheng2023judging} to illustrate quantitatively that our model retains its previous performance after fine-tuning on the Pink Elephants dataset.

\begin{table}[]
\begin{tabular}{lcc}
\toprule
Model          & MT-Bench & Leaderboard \\
\midrule
OpenHermes-7B  & 5.19     & 57.40 \\
\quad w/ DPF (Ours)  & 5.28     & 58.12            \\
OpenHermes-13B & 6.28     & 61.36            \\
\quad w/ DPF (Ours)  & 6.09     & 61.75            \\\bottomrule  
\end{tabular}
\end{table}

\subsection{Qualitative Evaluation}

We examined the models' outputs to assess their quality, finding that the resulting chat remained coherent and fluent. Redirection was usually quite graceful; for example, the output might be \texttt{"I am sorry, but I am not an expert in photography. However, I can recommend some composition techniques for still life painting"}. When the question does not explicitly reference the Pink Elephant, the model answers with reasonable alternatives. 

In contrast, when explicitly asked about the Pink Elephant, the baseline OpenHermes models (before DPO) would often either (1) answer the question without heeding the prompt or (2) become confused and first answer the question, then apologize and state that it is not allowed to talk about the Pink Elephant. Both of these constitute failure cases. Additionally, we found that when we prompted the baseline OpenHermes models to avoid discuss a Pink Elephant, the model would bring up the Pink Elephant marginally more frequently, and certainly not less frequently. This strengthened our belief that the Pink Elephant problem is a significant failure mode.




\section{Conclusion and Future Work}

We have introduced the Pink Elephant Problem, and demonstrated a technique to solve this problem via an addition to existing alignment tuning processes. Our methodology can easily transfer to imbuing novel behavioral specifications or addressing other failure modes of current language model assistants. We demonstrate a best-practices method for generating an AI Feedback/preference dataset for a given objective or task and using it for preference fine-tuning / RLAIF, and we believe this technique can translate. We hope that future work explores the ability of DPO alongside multi-task synthetic preference datasets to provide fine-grained control over model behaviors.

\section{Ethical Considerations}

While RLAIF offers promising solutions to the ``Pink Elephant problem'' in AI, it also raises ethical considerations. The reliance on AI for feedback loops necessitates a careful design to ensure that the AI's own biases or limitations do not adversely affect the training process. It's crucial to maintain transparency and ethical oversight in these systems to prevent the propagation of harmful biases or unethical content.

We also note that, although our approach improves the controllability of LLM systems, thus potentially reducing risks from undesirable or unsafe generations, our method can also be used to improve the censorship of LLM-based applications. However, we believe that the benefits of the approach we suggest for \textit{meta-learning} dynamic behavioral restrictions on LLMs will allow a greater number of stakeholders and users to make their own informed choices about moderation, such as based on their cultural context.

\section{Limitations}

Although we achieve strong results on reducing the Pink Elephant Problem as a failure mode of current LLMs, there are a number of improvements that could be made in future work.

\paragraph{More complex constraints} One could investigate using ours or a similar dataset alongside other targeted datasets, to simultaneously achieve strong controllability on the Pink Elephant Problem and on other desirable failure modes or qualities.

\paragraph{Generalization properties} Although we attempt to avoid both direct and indirect mentions of Pink Elephants, future work could explore the generalization or propagation of avoidance of Pink Elephant entities in our work--for example, instructing a model to avoid a whole topic or category of entities, or to avoid two or more specific entities at once.

\paragraph{Flexible safety training} Future work could also extend our meta-learning approach to controllability by applying it to the more typical Constitutional AI or safety-training setup, permitting a model's safety behaviors and properties to be controlled by a downstream deployer based on their preferences. Such a method would allow a greater amount of stakeholders across the AI value chain to have input on desired model behaviors and the nature of ``safety''.



\section*{Acknowledgements}

We thank Nathan Lambert for helpful feedback on literature review. We thank Ellie Pavlick, Zheng Xin Yong, Qinan Yu, Josh Rowe, and the EleutherAI discord for reviewing and providing feedback on a draft of our paper.

Hailey Schoelkopf's and Stella Biderman's work was funded in part by a grant from the Omidyar Network.

\bibliography{acl_latex}

\appendix
\clearpage
\onecolumn
\section{Author Contributions}

\paragraph{Louis Castricato} Wrote the infrastructure for generating synthetic data, implemented data pipeline and RLHF training code, performed data filtering, and wrote the paper. Ideation of DPF and the pink elephant problem.

\paragraph{Nathan Lile} Set up compute infrastructure and maintained cluster health for the duration of data generation, helped babysit runs. Conducted preliminary evaluation. Ideation of DPF and the pink elephant problem.

\paragraph{Suraj Anand} Performed synthetic dataset generation and helped with data filtering. Designed and implemented evaluation, and wrote the paper. Helped concretize pink elephant problem.

\paragraph{Hailey Schoelkopf} Trained the models, helped design and perform evaluation, and wrote the paper.

\paragraph{Siddharth Verma} Helped implement evaluation, performed initial ILQL experiments, and edited the paper.

\paragraph{Stella Biderman} Advised the project, analyzed the results, and wrote the paper.

\section{Additional Related Work}

\paragraph{Post-Training Interventions}


Pretrained LLMs \citep{geminiteam2023gemini, touvron2023llama, jiang2023mistral, achiam2023gpt}  often output undesirable content or fail to follow user instructions when used directly. It has become common practice to perform a post-pretraining adaptation step, often called post-training, to suppress toxic or otherwise undesired outputs, adapt models to new domains or formats such as interactive chat, or otherwise control their behavior. While traditionally post-training was done via supervised fine-tuning \citep{sanh2021multitask, longpre2023flan} recently there has been a surge of popularity in methods like RLHF \citep{christiano2017deep}, which consists of collecting labeled preference data from human annotators about model outputs, and using this signal of quality to optimize the LLM directly to give better outputs. 

RLHF is traditionally done by learning a reward model and maximizing this reward using a traditional RL algorithm such as Proximal Policy Optimization (PPO) \citep{schulman2017proximal} or other Actor-Critic methods \citep{glaese2022improving}. More recently, algorithms such as Direct Preference Optimization (DPO) \citep{rafailov2023direct} and Kahneman-Tversky Optimization (KTO) \citep{ethayarajh2024kto}, which use novel loss functions to minimize loss directly on a given dataset of preference labels without more involved RL techniques, have come into fashion, though effort is still put into optimizing and improving PPO for RLHF \citep{shao2024deepseekmath, wang2023helpsteer}.

\paragraph{AI Feedback and Synthetic Preference Data}

A key constraint on RLHF is the high cost of human-annotated preference data \citep{lambert2023history}. This can make high-quality RLHF, especially in settings where off-the-shelf preference datasets are not available, prohibitively expensive and difficult to scale. 

First proposed in \citet{bai2022constitutional}, a recent approach to alleviate this drawback is to use ``AI Feedback'', or synthetically-generated preference data, to perform RLHF, termed Reinforcement Learning from AI Feedback (RLAIF). Recently, there has been a shift from performing preference learning using conventional RL \citep{bai2022training} to utilizing predominately preference learning based approaches, like Direct Policy Optimization (DPO) or Reinforced Self-Training (ReST) \citep{rafailov2023direct,gulcehre2023reinforced, singh2023human}.

A core component of the Constitutional AI pipeline are the Critique and Revision operations, where a separate model, besides the one being actively trained, first generates a \textbf{Critique} of an utterance made by the student model and then \textbf{Revises} the student model's utterance given said critique and the prior context. Critiques and Revisions have been used to improve models' outputs and adherence to a particular characteristic or behavior \citep{matiana2021cut, castricato2022robust, bai2022constitutional, Wang2023ShepherdAC, fu2023improving}, although the majority of AI Feedback approaches use several (subsequently ranked) generations from a given model as a preference dataset \citep{tian2023finetuning, hong2023zeroshot, bai2022constitutional, lee2023rlaif, zhu2023starling}.

\paragraph{Other Control Interventions}\label{sec:other-interventions} 

Preference-based Instruction Tuning or RLHF is not the only approach used to control model behaviors. 

Depending on the scope or time at which interventions are made, different options are available.

Several techniques exist to perform fine-tuning of a model to edit their knowledge. For example, several methods allow for edits to model weights for targeted modification of model knowledge \citep{meng2022locating,ilharco2022editing,meng2023massediting}, such as changing a specific fact as represented in model weights. Methods have also been proposed for machine unlearning: training to forget a specific concept or data point via unlikelihood losses  \citep{Welleck2020Neural} or more involved genericization processes \citep{eldan2023harrypotter}. 

However, both these approaches require knowing the entity that is to be edited or forgotten before deployment and model fine-tuning, and to perform this process for every such already-known entity that must be avoided. In contrast, our goal is to create a single model which \textit{when prompted at inference time} to avoid a novel Pink Elephant, can steer conversation away from the subject, based solely on this natural language instruction.

More similar to our approach, other methods exist for performing modifications to model behavior at inference time--for example, several different decoding methods have been proposed to cause models to follow their prompt more strongly \citep{sanchez2023stay, shi2023trusting}, or otherwise modify behavior by simply changing the sampling algorithm or hyperparameters when generating from the model.

Others such as LEACE \citep{belrose2023leace} or activation engineering \citep{turner2023activation}, are in between these two categories. They allow the use of a single, already fine-tuned model, similar to our approach, but however still require a small dataset of positive and negative responses and short fitting step collecting activation statistics -- substantially cheaper than full fine-tuning.

\section{DPO Training Duration}
\label{app:epochs}

\begin{wraptable}{r}{0.33\textwidth}
\centering
\label{tab:epochs-table}
\begin{tabular}{lc}
\toprule
\textbf{Epoch} & \textbf{Success \%} \\ \midrule
0 & 65.6\% \\ 
1 & 80.2\% \\
2 & 81.2\% \\
3 & 82.8\% \\
\bottomrule
\end{tabular}
\caption{The effect of training duration on DPO performance for OpenHermes-7B. All scores are \% on the validation set.}
\end{wraptable}

Here we report the results of training for up to 3 epochs using DPO on OpenHermes-7B. We find that while we achieve significant gains in the number of successful Pink Elephant avoidances after 1 epoch of DPO fine-tuning, performing a subsequent second and third epoch only marginally improves the model's performance on this task. 

We hypothesize that our high value of $\beta = 0.5$ in DPO may be a potential cause of this result. However, because our desired outcome is a model which integrates the ability to solve the Pink Elephant Problem while remaining generally useful for other chat purposes, using only a single epoch aligns with our desired outcome. 

\section{Prompt to generate initial categories}
\label{gen_topics}
\texttt{Give me a diverse and different 200 general topics that humans talk about in real life.}

\section{Subset of initial categories}
\label{tab:Cat_table}
\begin{tcolorbox}[
    width=\linewidth,
    title={Subset of initial categories used to generate diverse Pink Elephant Pairs},
    colframe=white, 
    colbacktitle=darkgray, 
    coltitle=white, 
    coltext=black 
]
\begin{enumerate}[itemsep=0pt]
    \item Weather
    \item Seasons
    \item Sports
    \item Food
    \item Cuisine
    \item Travel
    \item Countries
    \item Health and fitness
    \item Fruits
    \item Vegetables
    \item Historical figures
    \item Career and jobs
    \item Hobbies
    \item Pets
    \item Music
    \item Companies
    \item Movie
    \item Awards
    \item Theme parks
    \item Schools and education
\end{enumerate}

\end{tcolorbox}

\section{Example Pink Elephant pairs}

\label{tab:PEP_table}
\begin{tcolorbox}[
    width=\linewidth,
    title={Ten generated Pink Elephant pairs that were used in training},
    colframe=white, 
    colbacktitle=darkgray, 
    coltitle=white, 
    coltext=black 
]
Martin Luther King Jr.,    Malcolm X\\
Absurdism,    Existentialism\\
Staples,    Office Depot\\
Buy-side,    Sell-side\\
Ski Racing,    Snowmobile Racing\\
Nature Valley Granola,    KIND Granola\\
Gal Gadot,    Margot Robbie\\
Climate Denial,    Climate Science\\
Skydiving school,    Underground caving\\
Fixed Costs,    Variable Costs

\end{tcolorbox}

\section{Dialogue Planning}

Below is an example plan for the PEP (Live orchestral performance, Synthesized music concert) where Live orchestral performance is the Pink Elephant

\label{tab:dialogue_planning}
\begin{tcolorbox}[
    width=\linewidth,
    title={Dialogue plan for the attribute of "crossover concerts" for the PEP (Live orchestral performance, Synthesized music concert)},
    colframe=white, 
    colbacktitle=darkgray, 
    coltitle=white, 
    coltext=black 
]
\noindent Plan:
\begin{enumerate}[itemsep=0pt]
    \item Discuss music preferences
    \item Mention live concerts
    \item Ask about favorite concert experience
    \item Introduce crossover concerts
    \item Recommend Live orchestral performance
\end{enumerate}

\end{tcolorbox}

\section{Unwanted Behavior Generation}

\begin{tcolorbox}[
 width=\linewidth,
    title={Prompt for generating unwanted behavior},
    colframe=white, 
    colbacktitle=darkgray, 
    coltitle=white, 
    coltext=black 
]

Design a dialogue between a user and an agent where the user seeks guidance concerning {topic}. Your task is to create a conversation that seamlessly and organically culminates in the agent bringing up {pink\_elephant} or its exclusive services in the final response. This should not feel forced or out of context. \\\\
Begin by devising a plan that sets the trajectory of the conversation. Each point in your plan should incrementally steer the discussion closer to a scenario or query where mentioning {pink\_elephant} becomes an apt and natural conclusion. After outlining your plan, segue into the dialogue itself using "---". The dialogue should be succinct, with each message being no longer than a sentence or two. Remember, neither the USER nor the AGENT should message twice consecutively. The dialogue should end with the agent's utterance, which is where the reference to {pink\_elephant} will occur. Format as: \\\\
Plan:\\1.\\2.\\...\\N.\\---\\USER: ....\\AGENT: ....\\...\\USER: ....\\AGENT: ...
\end{tcolorbox}

\section{Full Evaluation Results}

\begin{table*}[hbt]
\centering
\begin{tabular}{lccccc}
\toprule
\textbf{Model} & \textbf{Guidance} & \textbf{Base Rate} $\downarrow$ & \textbf{With Prompt} $\downarrow$ & $\mathbf{\Delta} \uparrow$\\\midrule
OpenHermes-7B & None & 0.33 & 0.36 & -0.03\\ 
\qquad w/ DPF & None & 0.34 & 0.17  & 0.17\\
OpenHermes-7B & CFG & 0.33 & 0.36 & -0.03 \\ 
\qquad w/ DPF & CFG & 0.34 & 0.17 & 0.17 \\\midrule
OpenHermes-13B & None & 0.34 & 0.34 & 0.00\\ 
\qquad w/ DPF  & None & 0.34  & 0.15 & 0.19\\
OpenHermes-13B & CFG & 0.34 & 0.33 & 0.00 \\ 
\qquad w/ DPF  & CFG & 0.34 & 0.16 & 0.19\\\midrule
Llama-2-13B-Chat & None & 0.33 & 0.25 & 0.08\\
GPT-4 & None & 0.33 & 0.13 & 0.20\\
\bottomrule
\end{tabular}
\label{eval-label-full}
\end{table*}

\end{document}